\pdfoutput=1

\documentclass[11pt,a4paper]{article}
\usepackage{acl}
\usepackage{latexsym}
\usepackage{breakurl}
\usepackage{latexsym}
\usepackage{csquotes}
\usepackage{times}
\usepackage[utf8]{inputenc}
\usepackage[T1]{fontenc}

\usepackage{footnote}

\usepackage{multirow}
\usepackage{afterpage}

\usepackage{xcolor}
\usepackage{colortbl}
\usepackage{tikz}
\definecolor{Gray}{gray}{0.9}
\definecolor{maroon}{cmyk}{0,0.87,0.68,0.32}

\def\defineCMYKcolor(#1,#2,#3,#4)#5{%
    \pgfmathsetmacro{\myc}{#1/255}%
    \pgfmathsetmacro{\mym}{#2/255}%
    \pgfmathsetmacro{\myy}{#3/255}%
    \pgfmathsetmacro{\myk}{#4/255}%
    \definecolor{#5}{cmyk}{\myc,\mym,\myy,\myk}%
}

\defineCMYKcolor(94,26,0,33){brightnavy}

\usepackage{microtype}
\usepackage{amsmath}
\usepackage{amsfonts}
\usepackage{amssymb}
\usepackage{graphicx}
\usepackage{cleveref}[2012/02/15]

\crefformat{footnote}{#2\footnotemark[#1]#3}

\makeatletter
\DeclareFontFamily{OMX}{MnSymbolE}{}
\DeclareSymbolFont{MnLargeSymbols}{OMX}{MnSymbolE}{m}{n}
\SetSymbolFont{MnLargeSymbols}{bold}{OMX}{MnSymbolE}{b}{n}
\DeclareFontShape{OMX}{MnSymbolE}{m}{n}{
    <-6>  MnSymbolE5
   <6-7>  MnSymbolE6
   <7-8>  MnSymbolE7
   <8-9>  MnSymbolE8
   <9-10> MnSymbolE9
  <10-12> MnSymbolE10
  <12->   MnSymbolE12
}{}
\DeclareFontShape{OMX}{MnSymbolE}{b}{n}{
    <-6>  MnSymbolE-Bold5
   <6-7>  MnSymbolE-Bold6
   <7-8>  MnSymbolE-Bold7
   <8-9>  MnSymbolE-Bold8
   <9-10> MnSymbolE-Bold9
  <10-12> MnSymbolE-Bold10
  <12->   MnSymbolE-Bold12
}{}

\let\llangle\@undefined
\let\rrangle\@undefined
\DeclareMathDelimiter{\llangle}{\mathopen}%
                     {MnLargeSymbols}{'164}{MnLargeSymbols}{'164}
\DeclareMathDelimiter{\rrangle}{\mathclose}%
                     {MnLargeSymbols}{'171}{MnLargeSymbols}{'171}
\makeatother
\PassOptionsToPackage{hyphens}{url}\usepackage{hyperref}
\newcommand{\ra}[1]{\renewcommand{\arraystretch}{#1}}
\newcommand{\ag}[1]{$\llangle$#1$\rrangle_{\vert q}$}
\def\aa{{\bf Text A }}
\def\bb{{\bf Text B }}

\title{The Fewer Splits are Better: Deconstructing Readability in Sentence Splitting}

\author{Tadashi Nomoto \\
  National Institute of Japanese Literature \\
  Tachikawa, Tokyo 190-0014, Japan \\
  \texttt{nomoto@acm.org} 
 \\}

\begin{document}
\maketitle
\begin{abstract}

In this work,  we focus on sentence splitting, a subfield of text simplification,  motivated  largely  by an unproven idea  that if you divide a sentence in pieces, it should become easier to understand.   Our primary goal in this paper is to find out whether this is true. In particular, we ask, does it matter whether we break a sentence into two or three?  We report on our findings based on Amazon Mechanical Turk. 

More specifically, we introduce a Bayesian modeling framework to  further investigate to what degree  a particular way of splitting the complex sentence affects readability, along with a number of other parameters adopted  from diverse perspectives, including clinical linguistics, and cognitive linguistics. The Bayesian modeling experiment provides clear evidence that bisecting the sentence leads to enhanced readability to a degree greater than what we create  by trisection. 

\end{abstract}

\section {Introduction}

\par In text simplification, one question people often fail to ask is, whether the technology they are driving truly helps people better understand texts. This curious indifference  may reflect the tacit recognition of the partiality of datasets covered by the studies \cite {xu_wei:2015} or  some murkiness that surrounds the goal of text simplification. 

\par As a way to address the situation, we examine a role of simplification in text readability, with a particular focus on sentence splitting.  The goal of sentence splitting  is  to  break  a sentence into small pieces in a way that they collectively preserve the original meaning.  A primary question we ask in this paper is, does a splitting of text affect readability? 
In the face of a large effort spent in the past on sentence splitting,  it comes as a surprise that none of the studies put this question directly to people; in most cases, they ended up asking whether generated texts \enquote*{looked simpler} than the original unmodified versions \cite{zhang-lapata-2017-sentence}, which of course does not say much about their readability. We are not even sure whether there was any agreement among people on what constituted simplification.

 \par  Another related question is,  how many pieces should we  break a sentence into?   Two, three, or more?  In the paper,  we  focus on a particular setting where we ask whether there is any difference in readability between two- and three-sentence splits.  We also report on how good or bad sentence splits are that are generated by a fine-tuned language model, compared to humans'.
 
 \par A general strategy we follow in the paper is to elicit judgments from people on whether simplification made a text anyway readable for them (Section~\ref{sec:analysis}), and do a Bayesian analysis of their responses to identify factors that may have influenced their decisions  (Section~\ref{sec:bayesian}).\footnote{We will make available on GitHub the data we created for the study soon after the paper's publication (they should be found under \url{https://github.com/tnomoto}).
 }


\section{Related Work}
Historically, there have been extensive efforts in  ESL (English as a Second Language) to explore the use of simplification as a way to improve reading performance of L2 (second language) students. \citet{crossley:2014} presented an array of evidence showing that simplifying text did  lead to an improved text comprehension by L2 learners as measured by  reading time and and accuracy of their responses to associated questions. They also noticed that simple texts had less lexical diversity, greater word overlap, greater semantic similarity among sentences than more complicated texts.  \citet{crossley:2011} argued for the importance of cohesiveness as a factor to influence the readability.  
Meanwhile, an elaborative modification of text was found to play a role in enhancing readability, which involves adding information to make the language less ambiguous and rhetorically more explicit. 
\citet{Ross1991SimplificationOE} reported that despite the fact that it made a text longer, the elaborative manipulation of a text  produced positive results, with L2 students scoring higher in comprehension questions on modified texts than on the original unmodified versions.

\par While  there have been concerted efforts in the past in the NLP community to develop  metrics and corpora purported to serve studies in simplification \cite{zhang-lapata-2017-sentence,sulem-etal-2018-bleu,narayan-etal-2017-split,botha-etal-2018-learning,niklaus-etal-2019-minwikisplit,kim-etal-2021-bisect,xu_wei:2015}, they fell far short of addressing how their work contributes to improving the text comprehensibility by readers.  Part of our goal is to break away from a prevailing view that  relegates the readability to a sideline.

\section{Method}\label{sec:method}


The data come from two sources, the Split and Rephrase Benchmark (v1.0) (SRB, henceforth) \cite{narayan-etal-2017-split} and WikiSplit \cite{botha-etal-2018-learning}. SRB consists of complex sentences aligned with a set of multi-sentence simplifications varying in size from two to four.  WikiSplit follows a similar format except that each complex sentence is accompanied only  by a two-sentence simiplification.\footnote{
We  used WikiSplit, together with part of SRB, exclusively to fine tune BART to give a single split (bipartite) simplification model, and SRB to develop test data to be administered to humans for linguistic assessments. SRB was derived from WebNLG \cite{gardent-etal-2017-creating} by making use of  RDFs associated with textual snippets to assemble simplifications.
}   We asked Amazon Mechanical Turk workers (Turkers, henceforth) to score simplifications on linguistic qualities as well as  to indicate whether they have any preference between two-sentence and three-sentence versions   in terms of readability.

\par We randomly sampled a portion of SRB, creating test data (call it $\cal H$), which consisted of  triplets of the form: $\langle S_0, A_0, B_0 \rangle$, $\dots$,  $\langle S_i, A_i, B_i \rangle$, $\dots$, $\langle S_m, A_m, B_m \rangle$, where $S_i$ is a complex sentence, $A_i$ a corresponding two-sentence simplification, and $B_i$ its three-sentence version. While $A$ alternates between versions created by BART and by human, $B$ deals only with manual simplifications.\footnote{
HSplit \cite{sulem-etal-2018-bleu} is another dataset (based on \citet{zhang-lapata-2017-sentence}) that gives multi-split simplifications. We did not adopt it here as the data came with only  359 sentences with limited variations in splitting.
} See  Table~\ref{tbl:test-set} for a further explanation. 

Separately, we extracted  from WikiSplit and SRB, another  dataset $\cal B$  consisting of complex sentences  as a source and  two-sentence simplifications as a target (Table~\ref{tbl:bart-data}) i.e. ${\cal B} =\{\langle S^\prime_0, A^\prime_0 \rangle$, $\dots$,  $\langle S^\prime_n, A^\prime_n \rangle\}$,  to use it to fine-tune a language model (BART-large).\footnote{\url{https://huggingface.co/facebook/bart-large}}
The fine-tuning was done using a code available at GitHub.\footnote{
\url{https://github.com/huggingface/transformers/blob/master/examples/pytorch/translation/run_translation.py}}
%
%

\begin{table}
\centering 
\ra{1.1}
\begin{tabular}{lcc}
& \sc bart & \sc hum \\
\sc a (two-sentence split) & 113 &  108 \\ 
\sc b (three-sentence split) & $-$  &  221\\
\end{tabular}
\caption{A break down of ${\cal H}$. 113 of them are of type A (bipartite split) generated by BART-large; 108 are of type A created by humans. There were 221 of type B (tripartite split), all of which were produced by humans. 
}\label{tbl:test-set}
\end{table}
\begin{table}
\centering\begin{tabular}{cc}
\sc train & \sc dev\\
1,135,009 (989,944) & 13,797(5,000) \\
\end{tabular}
\caption{A training setup for  BART. The data comes from SRB \cite{narayan-etal-2017-split} and WikiSplit \cite{botha-etal-2018-learning}. The parenthetical numbers indicate amounts of data that originate in WikiSplit \cite{botha-etal-2018-learning}.}\label{tbl:bart-data}
\end{table}
\begin{figure*}[t]

\includegraphics[width=17.3cm]{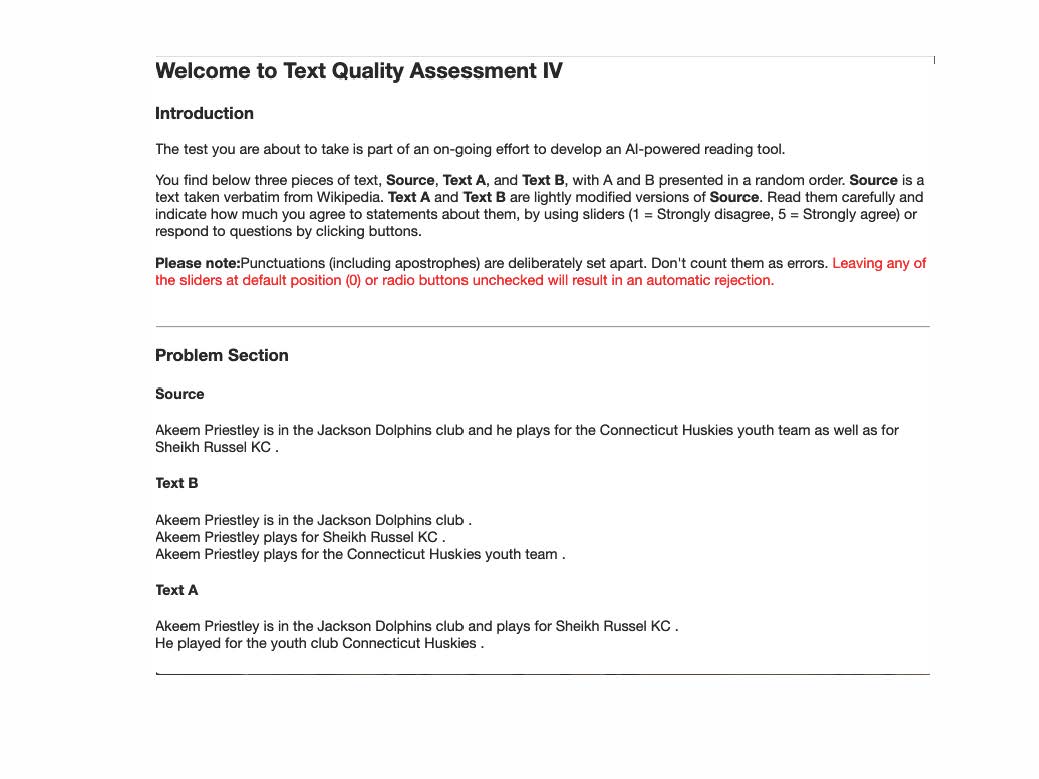}
\caption{A screen capture of HIT. This is what a Turker would be looking at when taking the test.} \label{fig:screen-capture}
\end{figure*}
\begin{table*}
\ra{1.1}
\centering\begin{tabular}{cccccc}
\multicolumn{1}{c}{\sc question}&\multicolumn{4}{c}{\sc available choices} &\\ \cline{2-5}
& \sc s  & \sc bart-a  & \sc hum-b  & \sc  not sure  & \sc total\\
\ag{\sc s,  bart-a} & 254 (0.32)& 527 (0.67) &--&  10 (0.01)& 791 \\
\ag{\sc s, hum-b}& 290 (0.37) & -- & 490 (0.62) & 11 (0.01)& 791\\
 & \sc s & \sc hum-a  & \sc hum-b  & \sc not sure&\sc total\\
\ag{\sc s, hum-a}& 253 (0.33) & 494 (0.65) &-- & 9  (0.01)& 756\\
\ag{\sc s, hum-b}& 288 (0.38) & -- & 463 (0.61)& 5 (0.01)& 756\\ 
\end{tabular}
%
\caption{Results from the Comparison Section. We are showing how many Turkers went with each available choice. S: source.  BART-A: BART-generated two-sentence simplification.  HUM-A: manual two-sentence simplification. HUM-B: manual three-sentence simplification. \ag{S, BART-A} asked  Turkers which of S and BART-A they found  easier to understand. 
67\% said  they would favor BART-A, and 32\% S, with 1\% not sure. \ag{S, HUM-B}  compares S and  HUM-B for readability.  \ag{S, HUM-A} looks at S and HUM-A.
} \label{tbl:results}
\end{table*}

\begin{table*}
\ra{1.2}
%
\centering\begin{tabular}{ccccc}
\multicolumn{1}{c}{\sc question}&\multicolumn{3}{c}{\sc available choices} &\\ \cline{2-4}
 &\sc bart-a  & \sc hum-b & \sc not sure& \sc total\\ 
\ag{\sc bart-a, hum-b} & 460 (0.58) & 316 (0.40)& 15 (0.02)& 791\\
& \sc hum-a & \sc hum-b & \sc not sure& \sc total\\
\ag{\sc hum-a, hum-b}  & 439 (0.58) & 301 (0.40)   & 16  (0.02) & 756\\ 
\end{tabular}
%
\caption{Comparison of two- vs three-sentence simplifications.  The majority went with two-sentence simplifications regardless of how they were generated.} \label{tbl:results-ab}
\end{table*}

\par  A task  (or a HIT in Amazon's parlance) we asked Turkers to do was to work on a three-part  language quiz. The initial  problem section  introduced a worker to three short texts, corresponding to a triplet $\langle S_i,  A_i, B_i \rangle$; the second  section asked about linguistic qualities of  $A_i$ and $B_i$ along three dimensions, {\it meaning}, {\it grammar}, and {\it fluency}; and in the third, we asked two comparison questions: (1) whether $A_i$ and $B_i$ are more readable than $S_i$, and (2) which of $A_i$ and $B_i$ is easier to understand.

Figure~\ref{fig:screen-capture} gives a screen capture of an initial section of the task.   Shown Under {\bf Source} is  a complex sentence  or $S_i$ for some $i$ .  \aa  and \bb correspond to $A_i$ and $B_i$, %
which were displayed in a random order.

In total, there were  221 HITs (Table~\ref{tbl:test-set}), each administered to seven people. All of the participants  were self-reported  native speakers of English with a degree from college or above. The participation was limited to residents in US, Canda, UK, Australia,  and New Zealand.

\section{Preliminary Analysis} \label{sec:analysis}

Table~\ref{tbl:results} summarizes results from comparison questions.  A question, labelled \ag{S, BART-A},  asks a Turker, which of Source and BART-A he or she finds  easier to understand, where BART-A is a BART generated two-sentence simplification.  We had  791 (113$\times$7)  responses, out of which 32\% said they preferred Source, 67\% liked BART better, and 1\% replied they were not sure.  Another question, labelled \ag{S, HUM-A}, compares Source to HUM-A, a two-sentence split  by human. It got 756 responses (108$\times$7). The result is generally parallel to \ag{S, BART-A}. 
The majority of people favored a two-sentence split  over a complex sentence.  
The fact that three sentence versions are also favored over complex sentences suggests that breaking up  a complex sentence improves readability, regardless of how many pieces it ends up with.

\begin{table}
\ra{1.1}
    \centering
    \begin{tabular}{ccc}
    category & \sc hum-a  & \sc  hum-b\\
        **fluency& 4.04 (0.39)& 3.75 (0.38) \\
         grammar &4.12 (0.32) &4.10 (0.32) \\
         meaning & 4.31 (0.36)& 4.33 (0.28)\\
    \end{tabular}
    \caption{Average scores and standard deviations for HUM-A and HUM-B. HUM-A is more fluent than HUM-B. Note: ** = $p < 0.01$.}
    \label{tbl:qual-5}
\end{table}
Table~\ref{tbl:results-ab} gives a tally of responses to comparison questions on two- and three-sentence splits.
More people voted for bipartite  over  tripartite simplifications. 
%
%
%
Tables~\ref{tbl:qual-5}  and \ref{tbl:qual-4} show scores on fluency, grammar, and meaning retention of simplifications,  comparing BART-A and HUM-B,\footnote{
As Tables~\ref{tbl:qual-5}  and \ref{tbl:qual-4} indicate, BART-A is generally comparable to HUM-A in the quality of its outputs, suggesting that what it generates is mostly indistinguishable from those by humans.
} on one hand,  and  HUM-A and HUM-S, on another,  on a scale of 1 (poor) to 5 (excellent).  In either case, we did not see much divergence between A and B in grammar and meaning, but they diverged the most in fluency. 
A T-test found the divergence statistically significant.  
Two-sentence simplifications generally scored higher on fluency (over 4.0) than three sentence counterparts (below 4.0). 
Table~\ref{tbl:examples} gives an example showing what  generated texts looked like in BART-A and HUM-A/B. 

\begin{table}[tbh]
\ra{1.1}
    \centering
    \begin{tabular}{ccc}
    category & \sc bart-a & \sc hum-b\\
         **fluency&4.04 (0.37) & 3.72 (0.36)\\
         grammar&4.07 (0.30)&4.05 (0.34)\\
         meaning&4.21 (0.38)&4.25 (0.35)\\
    \end{tabular}
    \caption{Average scores and standard deviations of BART-A and the corresponding HUM-B. BART-A is significantly more fluent than HUM-B. \enquote*{**} indicates the two groups are distinct at  the 0.01 level.}
    \label{tbl:qual-4}
\end{table}
\section{A Bayesian Perspective} \label{sec:bayesian}

\begin{table*}
\ra{.97}
\centering
%
%
\centering
\begin{tabular}{cp{.78\textwidth}}
\sc type & \hfil \sc text \hfil \null\\
\sc original & The Alderney Airport serves the island of Alderney and its 1st runway is surfaced with poaceae and has a 497 meters long runway .\\
\sc bart-a & {Alderney Airport serves the island of Alderney .}  The 1st runway at Aarney Airport is surfaced with poaceae and has 497 meters long .\\
\sc hum-a &The runway length of Alderney Airport is 497.0 and the 1st runway has a poaceae surface . The Alderney Airport serves Alderney .\\
\sc hum-b &The surface of the 1st runway at Alderney airport is poaceae . Alderney Airport has a runway length of 497.0 . The Alderney Airport serves Alderney .\\ 
\end{tabular}
\caption{Original vs. Modified}\label{tbl:examples}
\end{table*}

A question we are curious about at this point is what are the factors that led Turkers to decisions that they made. We  answer the question by way of building a Bayesian model  based on  predictors  assembled from   the past literature on readability and in related fields.

\subsection{Model}
We consider a Bayesian logistic regression.\footnote{
Equally useful in explaining relationships between potential causes and the outcome are Bayesian tree-based methods \cite{Chipman_2010,Linero2017ARO,nuti:2019}, which we do not explore here. The latter could become a viable choice when  an extensive non-linearity  exists between predictors and the outcome.}

\begin{equation}\label{eqn:model}
\begin{split} 
Y_j &\backsim Ber(\lambda), \\  
\text{logit}(\lambda) &=\beta_0 +  \sum_i^{m} \beta_i X_i,\\ 
\beta_i & \backsim {\cal N}(0, \sigma_i)\;\;  (0 \leq i \leq m) \\ 
\end{split}
\end{equation}
$Ber(\lambda)$ is a Bernoulli distribution with a parameter  $\lambda$. $\beta_i$ represents a coefficient tied to a random variable (predictor) $X_i$, where $\beta_0$ is an intercept. We assume that $\beta_i$,  including the intercept, follows a normal distribution with  the mean at $0$ and  the variance at $\sigma_i$. $Y_i$ takes either 1 or 0. $Y=1$ if a Turker finds  a two-sentence simplification more readable,  and $Y=0$ if a three-sentence version is preferred. 
\begin{savenotes}
\begin{table*}

\begin{tabular} {clp{.54\textwidth}c}
\sc category &\hfil \sc var name \hfil\null& \hfil \sc description \hfil\null& \sc value\\ \hline
\multirow{ 1}{*}{synthetic}&\bf bart & true if the simplification is generated by BART; false otherwise. &categorical\\ \hline
\multirow{15}{*}{cohesion} & {\bf ted1} & the tree edit distance (TED) between a source and its proposed simplification.\footnote{\label{fn:kassim}\url{https://github.com/jasonyux/FastKASSIM}} where TED represents the number of editing operations ({\it insert}, {\it delete}, {\it replace}) required to turn one parse tree into another; the greater the number,  the less the similarity \cite{Boghrati2018,zhang:edit:1989}.  & continuous\\
&{\bf ted2} &TED across sentences contained in the simplification.  & continuous\\%
&\bf subset & Subset based Tree Kernel \cite{collins-duffy-2002-new,moschitti-2006-making,chen2022fastkassim}\cref{fn:kassim} &continuous\\
&\bf subtree &{Subtree based Tree Kernel} \cite{collins-duffy-2002-new,moschitti-2006-making,chen2022fastkassim}\cref{fn:kassim} & continuous \\

&\bf overlap &Szymkiewicz-Simpson coefficient, a normalized cardinality of an intersection of two sets of words \cite{vijaymeena_etal:2016}.\footnote{\url{https://github.com/luozhouyang/python-string-similarity  
}} & continuous\\ \hline
\multirow{8}{*}{cognitive}&\bf frazier  & 
the distance from a terminal to  the root or the first ancestor that occurs leftmost \cite{frazier_1985}.
&continuous  \\
&\bf yngve &per-token count of non-terminals that occur to the right of a word in a derivation tree \cite{yngve:1960}. &continuous  \\
&\bf dep length & per-token count of dependencies in a parse \cite{magerman-1995-statistical,roark-etal-2007-syntactic}.&continuous \\
&\bf tnodes &  per-token count of nodes in a parse tree \cite{roark-etal-2007-syntactic}&continuous \\ 
\hline
\multirow{ 3}{*}{classic}&\bf dale & Dale-Chall readability score \cite{chall1995readability}\footnote{\label{fn:classic}\url{https://github.com/shivam5992/textstat}}&continuous \\
&\bf ease & Flesch Reading Ease \cite{flesch1979write}\cref{fn:classic}&continuous \\
&\bf fk grade &  Flesch-Kincaid Grade Level \cite{kincaid_et_al}\cref{fn:classic}&continuous \\ \hline
\multirow{3}{*}{perception}&\bf grammar & grammatical integrity (manually coded)&continuous \\
&\bf meaning & semantic fidelity (manually coded)&continuous \\  
&\bf fluency & language naturalness (manually coded)&continuous \\ \hline
\multirow{ 1}{*}{structural}&\bf split & true if the sentence is bisected; false otherwise.&categorical  \\ \hline
\multirow{ 1}{*}{informational}&\bf samsa & measures how much of the original content is preserved in the target \cite{sulem-etal-2018-semantic}.& continuous\\ 
\end{tabular}

\caption{Predictors \label{tbl:predictors}}
\end{table*}
\end{savenotes}

\subsection{Predictors}

We use predictors shown in Table~\ref{tbl:predictors}. They come in six categories: {\it synthetic}, {\it cohesion}, {\it cognitive}, {\it classic}, {\it perception} and {\it structural}. A {\it synthetic} feature  indicates whether the simplification was created with BART or not, taking {\it true} if it was and {\it false} otherwise. Those found  under {\it cohesion} are our adaptions of SYNSTRUT and  CRFCWO,  which are among the  diverse features \citet{mcnamara:2014} created to measure cohesion across sentences.  SYSTRUCT gauges the uniformity and consistency across sentences by looking at their syntactic similarities, or   by counting nodes in a common subgraph shared by neighboring sentences. We substituted SYSTRUCT  with {\bf tree edit distance} \cite{Boghrati2018}, as it allows us to handle multiple subgraphs,  in contrast to SYSTRUCT, which only looks  for a single common subgraph.  CRFCWO gives a normalized count of tokens found in common between two neighboring sentences. We emulated it  here with  the Szymkiewicz-Simpson coefficient, given as $O(X, Y) = \frac{\lvert X\cap Y\rvert}{\min (\lvert X \rvert, \lvert Y \rvert)}$.  

\par Predictors in the {\it cognitive} class are taken from works in clinical and cognitive linguistics \cite{roark-etal-2007-syntactic,Boghrati2018}.  They reflect various approaches to measuring the cognitive complexity  of a sentence. For example, 
 {\bf yngve} scoring defines a cognitive demand of a word as the number of non-terminals to its right in a derivation rule that are yet to be processed.   
 \subsubsection{ {\bf yngve} }
 \par
\begin{figure}[tbh]

\center\includegraphics[]{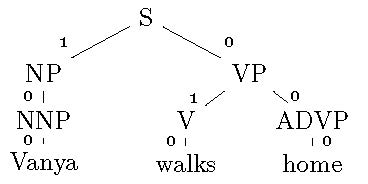}
\caption{Yngve scoring}\label{fig:yngve-scoring}
\end{figure}

 Consider Figure~\ref{fig:yngve-scoring}.  {\bf yngve}  gives every edge in the parse a number reflecting its cognitive cost.  NP gets \enquote*{1} because it has a sister node VP to its right. The cognitive cost  of a word is defined as the sum of numbers on a path from the root to the word.  In Figure~\ref{fig:yngve-scoring}, \enquote*{Vanya} would get $1+0+0=1$, whereas \enquote*{home} $0$. Averaging words' costs  gives us an Yngve complexity.
\subsubsection{\bf frazier}
\par
\begin{figure}[tbh]
\center\includegraphics[width=2.8cm]{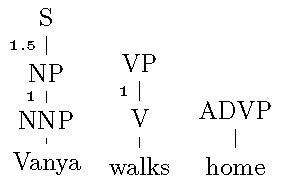}
\caption{Frazier scoring}\label{fig:frazier-scoring}
\end{figure}

  {\bf frazier} scoring views the syntactic depth of a word (the distance from a leaf to a first ancestor that occurs leftmost in a derivation rule) as a most important factor to determining the sentence complexity.   If we run {\bf frazier} on   the sentence  in Figure~\ref{fig:yngve-scoring},  it will get the score like one shown in Figure~\ref{fig:frazier-scoring}. \enquote*{Vanya} gets $1 + 1.5 = 2.5$, \enquote*{walks} $1$ and \enquote*{home} $0$ (which has no leftmost ancestor).
\citet{roark-etal-2007-syntactic} reported that both {\bf yngve} and {\bf frazier}  worked well in discriminating  subjects with mild memory impairement. 
\subsubsection{\bf dep length}
\par {\bf dep length} (dependency length) and {\bf tnodes} (tree nodes) are also among the features that  \citet{roark-etal-2007-syntactic}  found effective. The former measures the number of dependencies in a dependency parse, and the latter the number of nodes in a phrase structure tree.
\subsubsection{subset {\rm and} subtree}
 {\bf subset} and {\bf subtree} are both measures based on the idea of {\em Tree Kernel} \cite{collins-duffy-2002-new,moschitti-2006-making,chen2022fastkassim}.\footnote{
Tree Kernel  is a function defined as
$K(T_1, T_2)  =   \sum_{n_1 \in N (T_1)} \sum_{n_2 \in N (T_2)}  \Delta (n_1, n_2)$ where
 \[ \Delta (a, b) =\left \{
 \begin{array}{ll}
 0 &\mbox{if $a \neq b$};\\
 1 & \mbox{if $a = b$};\\
 \prod^{C(a)}_i (\sigma + \Delta(c^{(i)}_{a}, c^{(i)}_{b})) & \mbox{otherwise}.
 \end{array}\right.
 \] $C(a)$ = the number of children of $a$, $c^{(i)}_{a}$ represents the $i$-th child of $a$. We let $\sigma>0$. 
}
 The former considers how many subgraphs two parses share, while the latter how many subtrees. Note that subtrees are those structures that end with terminal nodes. 
\subsubsection{Classic readability features}
\par We also included features that have long been established in the readability literature as standard, i.e. Dale-Chall Readability, Flesch Reading Ease,  and Flesch-Kincaid Grade Level \cite{chall1995readability,flesch1979write,kincaid_et_al}.
\subsubsection{Perceptual  features}
\par Those found  in the {\it perception} category are from judgments Turkers made on the quality of simplifications we  asked them to evaluate. We did not provide any specific definition or instruction as to what constitutes grammaticality, meaning, and fluency during the task. So, it is most likely that their responses were spontaneous and perceptual.   
\subsubsection{split {\rm and} samsa}
Finally, we have {\bf split},  which records whether or not the simplification is bipartite:  it takes {\it true} if it is, and {\it false} if not.  {\bf samsa} is a recent addition to a battery of simplification metrics, which looks at how much of a propositional content in the source remains after a sentence is split \cite{sulem-etal-2018-semantic}. (The greater, the better.) 
We  standardized all of the features,  except for  {\bf bart} and {\bf split},  by turning them into  {\it z}-scores, where $ z = \frac{x - \bar{x}}{\sigma}$.


\subsection{Evaluation} \label{sec:evaluation}
We trained the model (Eqn.~\ref{eqn:model}) using {\sc bambi} \cite{bambi:2020},\footnote{\url{https://bambinos.github.io/bambi/main/index.html}} with the burn-in of 50,000 while making draws of  4,000,  on 4 MCMC chains (Hamiltonian).  As a way to  isolate the effect (or importance) of each predictor, we did two things: one was to look at a posterior distribution of each factor, i.e. a coefficient $\beta$  tied with a predictor,  and see how far it is removed from 0; another was to conduct an ablation study where we looked at how the absence of a feature affected the model's performance,  which we measured with a metric known as  \enquote*{Watanabe-Akaike Information Criterion} (WAIC) \cite{Watanabe10asymptoticequivalence,Vehtari_2016}, a Bayesian incarnation of AIC \cite{burnham2003model}.\footnote{
WAIC is given as follows.
\begin{equation}
\text{WAIC}=\sum_i^n \log  \mathbb{E}  [p(y_i \vert \theta)] - \sum_i^n \mathbb{V} [\log p(y_i \vert \theta)].
\end{equation}
$\mathbb{E}  [p(y_i \vert \theta)]$ represents the average likelihood under the posterior distribution of $\theta$, and  $\mathbb{V} [\alpha]$ represents the sample variance of $\alpha$,  i.e. $\mathbb{V}[\alpha] = \frac{1}{S-1}\sum_1^S (\alpha_s - \bar{\alpha})$, where $\alpha_s$ is a sample draw from $p(\alpha)$.  A higher WAIC score  indicates a better model.  $n$ is the number of data points.
}
\begin{figure}
\hspace{-.0cm}\includegraphics[width=7.8cm]{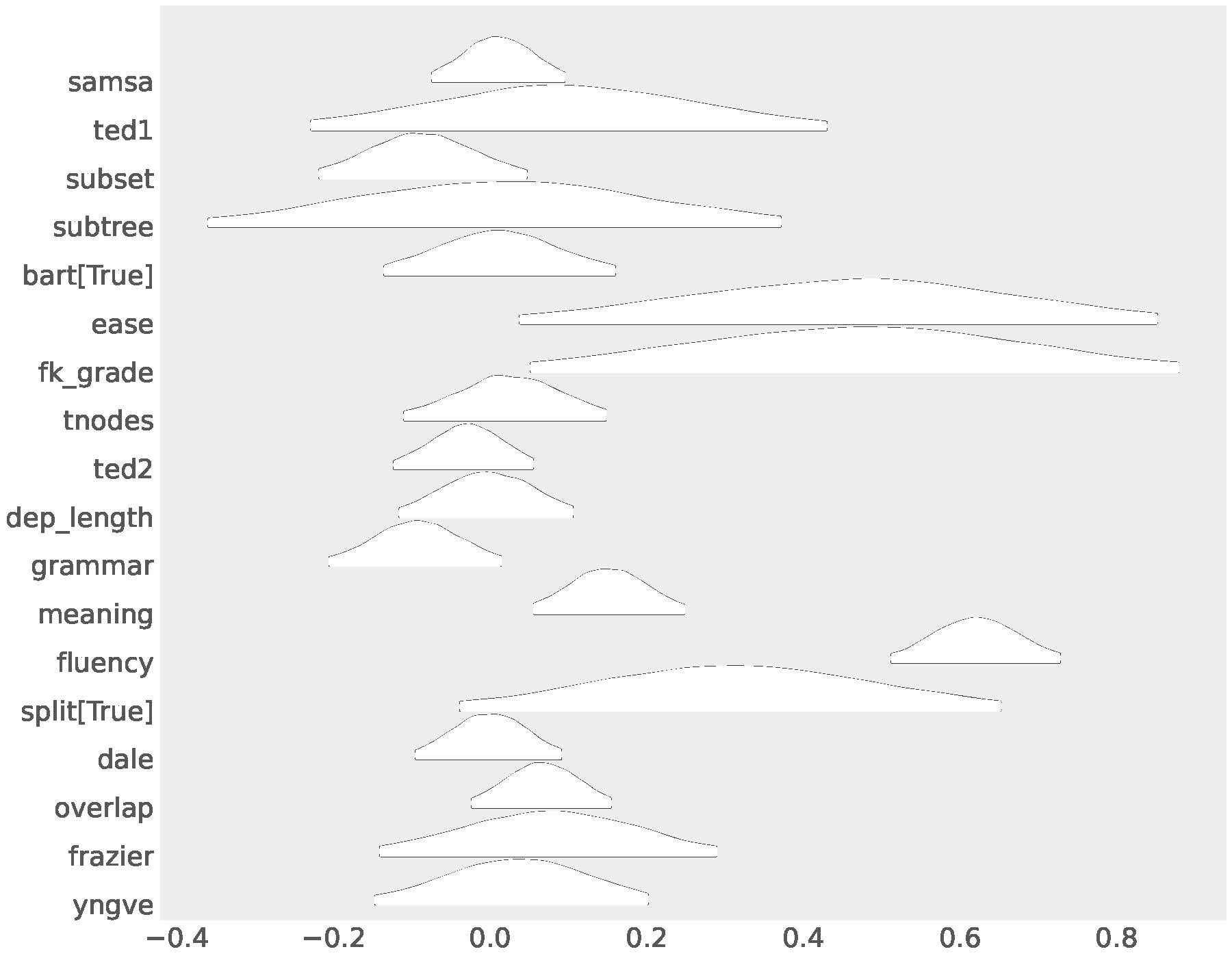}
\caption{Posterior distributions of coefficients ($\beta$'s)  in the full model. The further the distribution moves away from 0, the more relevant it becomes to predicting the outcome. \label{fig:post}}
\end{figure}
\par Figure~\ref{fig:post} shows what posterior distributions of parameters associated with predictors looked like after 4,000 draw iterations with MCMC.  None of the chains associated with the parameters exhibited divergence.  We achieved $\hat{R}$ {between 1.0 and 1.02,} for all $\beta_i$,  a fairly solid stability \cite{gelman-rubin:1992}, indicating  that  all the relevant parameters had successfully converged.\footnote{
$\hat{R} = \text{the ratio of within- and between-chain variances}$, a standard tool to check for convergence \cite{,lambert:2018}.  The closer the ratio is to the unity, the more likely MCMC chains have converged.  }
\par At a first glance, it is a bit challenging what  to make of Figure~\ref{fig:post},  but a generally accepted rule of thumb is to assume distributions that center around 0 as of less importance in terms of explaining observations,  than those that appear away from zero.  If we go along with the rule, then the most likely candidates that affected readability are: {\bf ease}, {\bf subset}, {\bf fk grade}, {\bf grammar}, {\bf meaning}, {\bf fluency}, {\bf split},  and {\bf overlap}. What remains unclear is, to what degree the predictors affected readability.
\par One good way to find out  is to do an ablation study,  a method to isolate the effects of an individual factor by examining how seriously its removal from a model degrades its performance.  The result  of the study is shown in Table~\ref{tbl:main_waic}.  Each row represents performance in WAIC of a model with a particular predictor  removed.  Thus, \enquote*{ted1} in Table~\ref{tbl:main_waic} represents a model  that includes all the predictors in Table~\ref{tbl:predictors}, except for {\bf ted1}. A row in blue represents a full model which had none of the features disabled. Appearing above the base model means that a removal of a feature had a positive effect, i.e. the feature is redundant. Appearing below means that the removal had a negative effect, indicating that we should not forgo the feature.  A feature becomes more relevant as we go down, and becomes less relevant as we go up the table. Thus the most relevant is {\bf fluency}, followed by {\bf meaning},  the least relevant is {\bf subtree}, followed by {\bf dale}, and so forth.  We can tell from Table~\ref{tbl:main_waic}  what predictors we need to keep to explain the readability: they are {\bf grammar}, {\bf split}, {\bf fk grade}, {\bf ease}, {\bf meaning} and {\bf fluency} (call them \enquote*{select features}). Note that {\bf bart} is in the negative realm, meaning that from a perspective of readability, people did not care about whether the simplification was done by human or machine.  {\bf samsa} was also found in the negative domain, implying that for a perspective of information, a two-sentence splitting carries  just as much information as  a three way division of a sentence.%
\par
\begin{figure}
\hspace{-.0cm}\includegraphics[width=7.8cm]{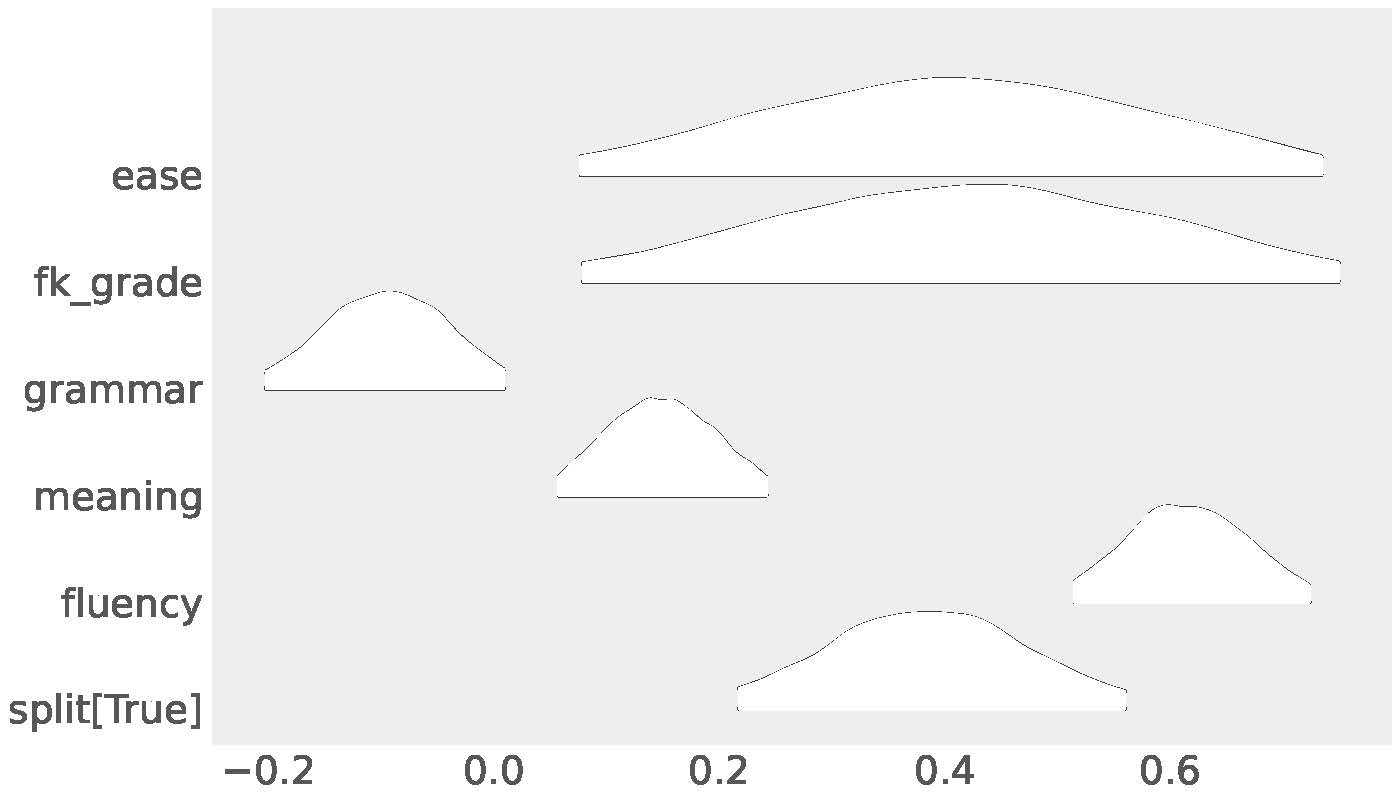}
\caption{Posterior distributions of the coefficient parameters in the reduced model.\label{fig:bestpost}}
\end{figure}

 To further nail down to what extent they are important, we ran another ablation experiment involving the select features alone. The result is shown in Table~\ref{tbl:best_waic}.  At the bottom is {\bf fluency}, the second to the bottom is {\bf split}, followed by {\bf meaning}, and so forth.
 As we go up the table, a feature  becomes less and less important. 
The posterior distributions of these features are shown in Figure~\ref{fig:bestpost}.\footnote{We found that they had $1.0 \leq \hat{R} \leq 1.01$, a near-perfect stability. Settings for MCMC, i.e. the number of burn-ins and that of draws,  were set to the same as before.} 
Not surprisingly, they are found away from zero, with {\bf fluency} furtherest away. The result indicates that contrary to the popular wisdom  that classic readability metrics such as {\bf ease}, and {\bf fk grade}, are of little use,
they  had a large sway on decisions people made when they were asked about readability. 

\begin{table*}
\ra{1.2}
\begin{tabular}{clrrrrrr}
 \it effect&  predictor & rank$\uparrow$  & waic$\uparrow$ &  p\_waic$\downarrow$& d\_waic$\downarrow$ &  se$\downarrow$ &  dse$\downarrow$ \\
\multirow{12}{*}{$-$}
&subtree & 0 &-1899.249&17.797&0.000&17.787&0.000\\
&dale & 1 &-1899.287&17.852&0.038&17.791&0.207\\
&dep\_length & 2 &-1899.362&17.916&0.113&17.777&0.211\\
&yngve & 3 &-1899.406&17.904&0.157&17.777&0.464\\
&tnodes & 4 &-1899.414&17.898&0.165&17.797&0.408\\
&bart & 5 &-1899.421&17.967&0.172&17.786&0.216\\
&samsa & 6 &-1899.450&18.018&0.201&17.776&0.315\\
&ted1 & 7 &-1899.557&17.996&0.308&17.771&0.575\\
&ted2 & 8 &-1899.632&18.019&0.383&17.782&0.624\\
&frazier & 9 &-1899.740&18.096&0.492&17.779&0.708\\
&subset & 10 &-1900.069&17.811&0.820&17.741&1.282\\
&overlap & 11 &-1900.431&17.966&1.182&17.750&1.511\\
\multirow{1}{*}{\color {blue} \it ref.}
&\color {blue} base & \color {blue} 12 &\color {blue} -1900.532&\color {blue} 19.089&\color {blue} 1.283&\color {blue} 17.787&\color {blue} 0.208\\ 
\multirow{6}{*}{$+$}
&grammar & 13 &-1900.780&17.979&1.531&17.698&1.657\\
&split & 14 &-1900.852&18.030&1.603&17.697&1.776\\
&ease & 15 &-1901.657&17.962&2.408&17.670&2.064\\
&fk\_grade & 16 &-1901.710&18.030&2.462&17.685&2.049\\
&meaning & 17 &-1903.795&17.885&4.546&17.425&3.071\\
&fluency & 18 &-1965.386&17.938&66.137&14.067&11.349\\
\end{tabular}

\caption{Comparison in WAIC. {\it p\_waic} = the  effective number of parameters \cite{spiegelhalter2002bayesian}, a measure to estimate the complexity of the model: the greater, the more complex. {\it d\_waic} = the distance in WAIC to the top model. {\it se} = standard error of WAIC estimates. {\it dse} =  standard error of differences in WAIC estimates between the top model and each of the rest.  $\uparrow$ means that higher is better. $\downarrow$ indicates the opposite.\label{tbl:main_waic}}
\end{table*}
\begin{table*}
\ra{1.2}
\begin{tabular}{lrrrrrr}
predictor & rank$\uparrow$  & waic$\uparrow$ &  p\_waic$\downarrow$& d\_waic$\downarrow$ &  se$\downarrow$ &  dse$\downarrow$ \\ 
\color {blue} base & \color {blue} 0
&\color {blue} -1891.901&\color {blue} 7.181&\color {blue} 0.000&\color {blue} 17.485&\color {blue} 0.000\\ 
grammar & 1
&-1892.235&6.183&0.335&17.365&1.672\\
ease & 2
&-1893.515&6.137&1.614&17.350&2.324\\
fk\_grade & 3
&-1893.626&6.161&1.726&17.366&2.358\\
meaning & 4
&-1895.308&6.145&3.407&17.111&3.059\\
 split & 5
&-1900.028&6.169&8.127&17.038&4.247\\
fluency & 6
&-1956.041&5.935&64.140&13.784&11.289\\
\end{tabular}

\caption{Results in WAIC for the reduced model\label{tbl:best_waic}}
\end{table*}

\section{Conclusions}
In this work, we asked two questions:  does cutting up a sentence help the reader better understand the text?  and if so, does it matter how many pieces we break it into?  We  found that splitting does allow the reader to better interact with the text (Table~\ref{tbl:results}) and moreover, two-sentence simplifications are clearly favored over three-sentence simplifications (Tables~\ref{tbl:results},\ref{tbl:main_waic},\ref{tbl:best_waic}). Why two-sentence splits make a better simplification is something of a mystery.  A possible answer may lie in a potential disruption splitting may have caused in a sentence-level discourse structure, whose integrity \citet{crossley:2011,crossley:2014} argued, constitutes a critical part of simplification,
a topic that we believe is worth  a further exploration in the future.

\section{Limitations}

\begin{itemize}
 \item We did not consider cases where a sentence is split into more than three. This is mainly due to our failure to find a dataset containing manual simplifications of length greater than three in a large number. While it is unlikely that our claim in this work does not hold for cases beyond three, testing the hypothesis on cases that involve more than three sentences would be desirable. 
 
 \item 
 A cohort of people we solicited for the current work are generally well educated adults who speak English as the first language. Therefore, the results we found in this work may not necessarily hold for L2-learners, minors, or  those who do not have college level education. 
 
 \end{itemize}
\section{Acknowledgement}
We thank anonymous reviewers for sharing  with us  their comments and ideas. We note  their effort with much gratitude and appreciation.

\bibliographystyle{acl_natbib}       
\bibliography{ling1-u}

%

\end{document}